\documentclass{article}

    \PassOptionsToPackage{sort, square, numbers}{natbib}

\usepackage[preprint]{neurips_data_2024}
\usepackage{graphicx}
\usepackage{float}
\usepackage[table,xcdraw]{xcolor}
\usepackage{siunitx}
\usepackage{subcaption}
\usepackage{multirow}
\usepackage{caption} 
\usepackage{wrapfig}
\usepackage[utf8]{inputenc} 
\usepackage[T1]{fontenc}    
\usepackage{hyperref}       
\usepackage{url}            
\usepackage{booktabs}       
\usepackage{amsfonts}       
\usepackage{nicefrac}       
\usepackage{microtype}      
\usepackage{xcolor}         

\newcommand{\dataset}[1]{\textsc{IndicVoices-R}}
\title{\dataset{}: Unlocking a Massive Multilingual Multi-speaker Speech Corpus for Scaling Indian TTS}

\author{%
  Ashwin Sankar\textsuperscript{\textdagger} \And Srija Anand\textsuperscript{\textdagger} \And Praveen Srinivasa Varadhan\textsuperscript{\textdagger} \And Sherry Thomas \And Mehak Singal \And  Shridhar Kumar \And Deovrat Mehendale \And Aditi Krishana \And Giri Raju \And Mitesh Khapra  \\\\
  AI4Bharat, Department of Computer Science \& Engineering\\
  Indian Institute of Technology Madras\\
  Chennai, Tamil Nadu, India - 600036 \\
  \textsuperscript{\textdagger} Equal contribution
}

\begin{document}

\definecolor{amber(sae/ece)}{rgb}{1.0, 0.49, 0.0}

\newcommand{\MK}[1]{\textcolor{blue}{#1}}
\newcommand{\SB}[1]{\textcolor{orange}{#1}}
\newcommand{\AG}[1]{\textcolor{purple}{#1}}

\definecolor{cerise}{rgb}{0.87, 0.19, 0.39}
\newcommand{\comment}[1]{\textcolor{cerise}{#1}}

\maketitle

\begin{abstract}
\label{sec:abstract}
Recent advancements in text-to-speech (TTS) synthesis show that large-scale models trained with extensive web data produce highly natural-sounding output. However, such data is scarce for Indian languages due to the lack of high-quality, manually subtitled data on platforms like LibriVox or YouTube. To address this gap, we enhance existing large-scale ASR datasets containing natural conversations collected in low-quality environments to generate high-quality TTS training data. Our pipeline leverages the cross-lingual generalization of denoising and speech enhancement models trained on English and applied to Indian languages. This results in IndicVoices-R (IV-R), the largest multilingual Indian TTS dataset derived from an ASR dataset, with 1,704 hours of high-quality speech from 10,496 speakers across 22 Indian languages. IV-R matches the quality of gold-standard TTS datasets like LJSpeech, LibriTTS, and IndicTTS. We also introduce the IV-R Benchmark, the first to assess zero-shot, few-shot, and many-shot speaker generalization capabilities of TTS models on Indian voices, ensuring diversity in age, gender, and style. We demonstrate that fine-tuning an English pre-trained model on a combined dataset of high-quality IndicTTS and our IV-R dataset results in better zero-shot speaker generalization compared to fine-tuning on the IndicTTS dataset alone. Further, our evaluation reveals limited zero-shot generalization for Indian voices in TTS models trained on prior datasets, which we improve by fine-tuning the model on our data containing diverse set of speakers across language families. We open-source all data and code, releasing the first TTS model for all 22 official Indian languages.

\textbf{Dataset Link -} \url{https://github.com/AI4Bharat/IndicVoices-R}

\end{abstract}

\section{Introduction}
\label{sec:Intro}
Scaling training data has been crucial in achieving better zero-shot speaker and style generalization for English Text-To-Speech (TTS) systems, with studies demonstrating the benefits of increasing data from a few hundred hours to beyond 10,000 hours \cite{kharitonov2023spear, shen2018natural, ju2024naturalspeech, peng2024voicecraft, wang2023valle}. Indeed, state-of-the-art models like Natural Speech 3 \cite{ju2024naturalspeech} are trained on 200,000 hours of data. Additionally, using conversational or extempore speech has been shown to enhance the naturalness of TTS systems compared to read-speech data. However, scaling training for Indian TTS has been challenging due to limited resources. Existing TTS datasets for Indian languages collectively cover only up to 14 of the 22 scheduled languages of India, with 1-2 speakers per language and primarily consisting of read-speech recordings which are not rich in prosody and expression. These limitations significantly hinder the progress of Indian TTS towards natural-sounding speech. 

One common approach to scale the training data is to mine TTS data from online public sources such as YouTube. However given the low resource status of Indian languages, finding manually transcribed data for all 22 Indian languages on the internet are challenging, and the available data often lacks professional studio quality. Further, such in-the-wild speech data typically contains overlapping speakers, music, and a low signal-to-noise ratio. Instead, we choose to repurpose existing ASR datasets for TTS, a popular approach that has unlocked better data in English, such as with LibriTTS-R \cite{koizumi2023libritts} and LibriLight (60k hours) \cite{kahn_2020}. For doing so, several candidate Indian ASR datasets exist, including KathBath\cite{javed2023indicsuperb}, Shrutilipi \cite{effectiveness2023bhogale}, FLEURS \cite{conneau2022fleurs}, and IndicVoices\cite{javed2024indicvoices}. Among these, IndicVoices is the most promising due to its unique characteristics: it covers all 22 scheduled languages of India, includes both read-speech and conversational formats, and involves a large number of speakers. These characteristics are also desirable in a TTS dataset, enabling the creation of human-like speech across all scheduled Indian languages while achieving zero-shot speaker generalization.

\begin{wrapfigure}{r}{0.45\textwidth}
    \centering
    \includegraphics[width=0.39\textwidth]{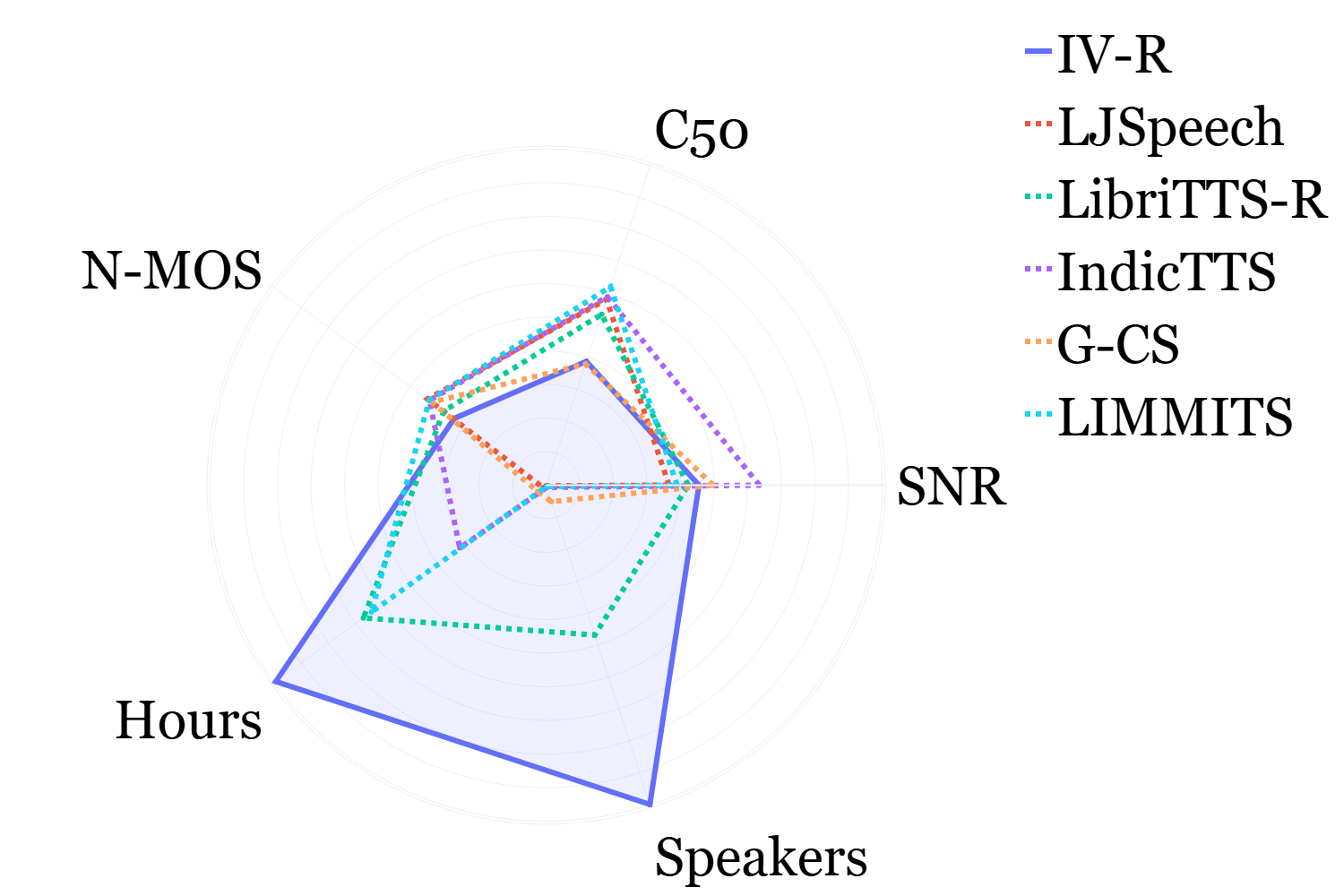}
    \caption{\small IndicVoices-R (IV-R) surpasses existing TTS datasets in terms of sheer volume of hours and speakers while having comparable speech quality to such TTS datasets, as assessed across a range of speech-quality metrics.}
    \label{fig:main}
\end{wrapfigure}

We propose a pipeline to denoise and enhance ASR data, effectively addressing the poor quality of audio samples, which often include background noise, intermittent chatter, echoes, and high reverb. Such data is frequently collected at a lower sampling rate of 16 KHz, compared to the 48 KHz typically found in professional studio recordings. We leverage the significant progress in Speech Enhancement and Restoration models to effectively address these issues. We find that these models, although pre-trained on English, show good cross-lingual generalization and can enhance Indian speech data without affecting intelligibility. Using the proposed pipelined we build IndicVoices-R (IV-R), the largest multilingual Indian Text-to-Speech (TTS) dataset derived from IndicVoices, unlocking 1,704 hours of high-quality speech data from 10,496 speakers across 22 Indian languages. As seen in Figure \ref{fig:main}, our dataset matches the speech and sound quality of gold-standard TTS datasets like LJSpeech, LibriTTS, and IndicTTS, as measured by the NORESQA \cite{noresqamos}, SNR, and C50 scores \cite{lavechin2023brouhaha}. We complement IV-R, by also releasing a comprehensive benchmark to test zero-shot, few-shot, and many-shot speaker generalization of Indian TTS systems.

Using IV-R, we enhance the zero-shot speaker generalization of an English pre-trained VoiceCraft model and also enables it to support multiple Indian speakers and languages. We show the former by contrasting its performance with a model fine-tuned on a less diverse Indian TTS dataset. Along with all our data and code, the first TTS model that supports all 22 scheduled Indian languages will be publicly released as part of this work.

\section{Background}
\label{sec:related-work}
\textbf{Achieving Human-level Speech Quality} NaturalSpeech \cite{tan2024naturalspeech} is a pioneering work that achieved human-level speech quality in neural TTS with large-scale phoneme encoder pre-training and improved architectures leveraging fully end-to-end training. Similarly, StyleTTS2 \cite{li2023styletts2} surpasses the quality of human recordings by using multiple techniques such as style diffusion and adversarial training with large speech language models such as WavLM \cite{chen2022wavlm}.   However, a limitation of these models trained on professional studio-quality datasets is that they often contain only single-speaker data or cover limited speaking styles such as read-speech or narration, thereby limiting the ability of these TTS systems to generalize to a variety of voices, speaking styles, and prosodic variations.  To overcome this barrier, several works \cite{kharitonov2023spear, ju2024naturalspeech, peng2024voicecraft, wang2023valle, shen2023naturalspeech2} scale TTS training using up to 1B parameters on large-scale multi-speaker found data consisting upto 100K hours.

\textbf{Speech-Text Datasets}
LJSpeech \cite{ljspeech17} is a foundational TTS dataset that releases 24 hours of single-speaker recordings from passages of non-fiction books. LibriTTS \cite{zen2019libritts} releases multi-speaker data derived from LibriVox audiobooks, providing 585 hours of speech from 2,456 speakers. VCTK \cite{veaux2017cstr} offers another significant resource with approximately 44 hours of recordings from 109 native English speakers with various accents. 
The introduction of several multi-speaker datasets \cite{zen2019libritts, veaux2017cstr, shi2020aishell3} has enabled the development of more prosodic TTS adaptive to multiple speaking styles and speaker voices. More recently, there has been a growing demand for large (un)labeled speech corpora \cite{Pratap2020MLSAL, chen2021gigaspeech} for training TTS systems. Existing Indian speech-text datasets like IndicTTS \cite{baby2016resources}, IndicSpeech \cite{srivastava2020IndicSpeech}, Google-CrowdSourced (Google-CS) \cite{he2020open} and LIMMITS \cite{googleLIMMITS24} are either considerably smaller in scale, exhibit lower speaker or style diversity compared to open-source multilingual datasets. This limited scale restricts the development of robust and versatile TTS models for Indian languages. In this work, we take inspiration from LibriTTS-R \cite{koizumi2023libritts} dataset, and attempt to enhance and restore IndicVoices \cite{javed2024indicvoices} to bridge the data gap of large-scale diverse multi-speaker data for Indian languages.

\textbf{TTS for Indian Languages} Initial neural works \cite{prakash2019building, prakash2020generic} that build TTS for Indian languages explore the importance of pooling speaker data across multiple languages within the same language family to overcome the data constraints and train good quality TTS systems. Subsequent work \cite{Kumar2022Towards} outlines design choices for training Indian TTS and releasing better quality monolingual multi-speaker models for 13 Indian languages. More recently, work in  \cite{prakash2023exploring} demonstrates that training multilingual, multi-speaker text-to-speech systems based on language families, like Indo-Aryan and Dravidian, can effectively leverage limited transcribed data and adapt to new languages in low-resource scenarios. 

\textbf{TTS Benchmarks} Comparison of state-of-the-art TTS has often been challenging due to the lack of standardized benchmarks. Conventionally, a couple of works \cite{ping2017deep, ren2019fastspeech, shen2018natural} evaluate on an agreed-upon set of 50 particularly hard sentences. Several works evaluate on the train-test splits of the original dataset they train on, especially for works training on LJSpeech and LibriTTS. EXPRESSO \cite{nyugen2023expresso} offers 47 hours of North American English data from 4 speakers across 26 styles, providing a benchmark for TTS prosodic and speaker-style variation. BaseTTS \cite{lajszczak2024base} introduces benchmarks for questions, emotions, compound nouns, syntactic complexity, foreign words, punctuations, and paralinguistics. However, the Indian TTS community faces a significant gap in the availability of benchmarks for TTS evaluation. We aim to address this by releasing the first zero-shot, few-shot, and many-shot cross speaker benchmark, for all 22 Indian languages.

\section{\dataset{}}
\label{sec:IndicVoices-R}
\dataset{} is derived from IndicVoices, an existing large-scale multilingual multi-speaker speech corpus for Indian languages. This section explains our rationale behind choosing IndicVoices, and details our data pipeline to restore this dataset to TTS quality. We also compare our enhanced version, IndicVoices-R, with several existing TTS datasets and show that it achieves on-par quality. 
Finally, we provide detailed statistics and the format of our released dataset. IndicVoices-R, boasts several key features that make it an ideal dataset choice for scaling Indian TTS systems:  

\textbf{(i) Comprehensive Coverage:} It is the first dataset to encompass all 22 Indian languages, offering between 9 to 175 hours of speech data per language. 

\textbf{(ii) Speaker Diversity:} With a vast pool of over 10,496 speakers (greater than any existing TTS dataset) representing various demographics and linguistic backgrounds, it ensures rich diversity crucial for attaining good cross-speaker generalization in TTS. 

\textbf{(iii) Natural Recordings:} The dataset predominantly consists of extempore recordings (93.25\%), capturing spontaneous speech, which is pivotal for achieving naturalness in synthesized speech. 

\textbf{(iv) High-Quality Samples:} The quality of the dataset's samples matches or exceeds that of several existing large-scale multi-speaker TTS datasets, underscoring its efficacy in TTS model training.

\begin{figure}[!t]
    \includegraphics[width=\linewidth]{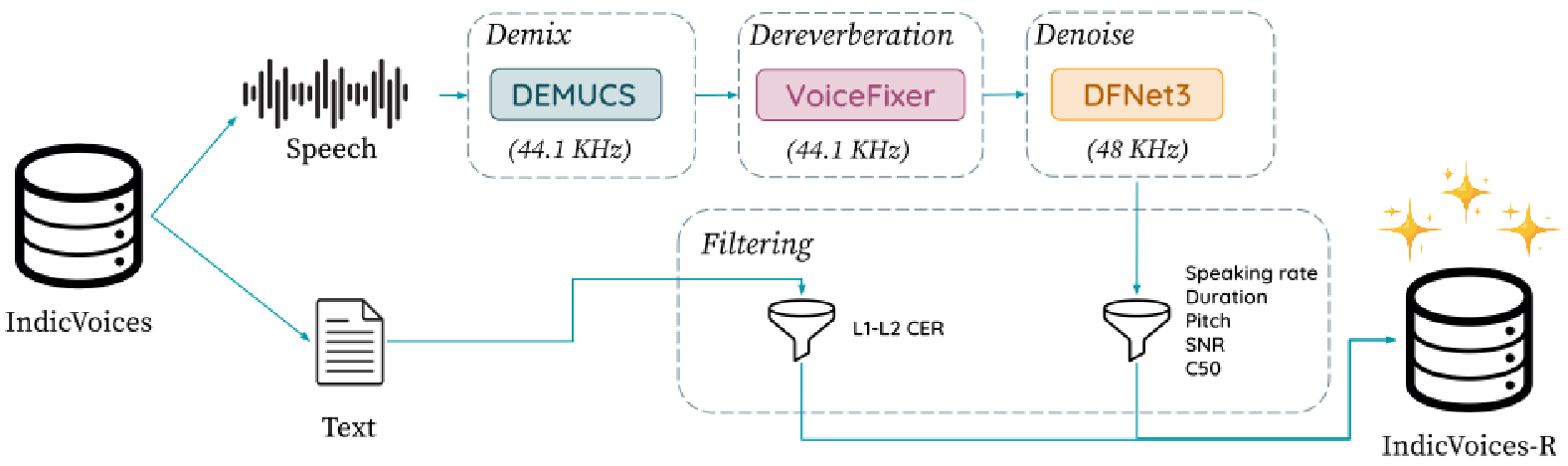}
    \caption{Data pipeline for restoring IndicVoices that demixes, dereverberates, and denoises speech samples, followed by filtering of speech-text pairs to yield an enhanced IndicVoices-R dataset.}
    \label{fig:data-pipeline}
\end{figure}

\subsection{Choice of ASR Dataset} 
While several Automatic Speech Recognition (ASR) datasets like Kathbath \cite{superb2023javed}, Shrutilipi \cite{effectiveness2023bhogale}, IndicVoices \cite{javed2024indicvoices}, and Dhwani \cite{javed2021dhwani} exist, IndicVoices emerges as the most suitable choice for constructing an Indian Text-to-Speech (TTS) dataset. Particularly, we would like to highlight several key advantages of IndicVoices based on (i) data quality, (ii) speaker and language coverage, (iii) diversity, and (iv) ethical considerations. Unlike the other datasets, which are limited to lower sampling rates (\SI{16}{\kilo\hertz}), IndicVoices prioritizes high-fidelity recordings at \SI{44.1}{\kilo\hertz}. Furthermore, manual annotation with a stringent quality assurance process guarantees high-quality text alignments with audio samples. This surpasses the need for relying on ASR-based transcription that would potentially introduce errors when labelling other large-scale speech corpuses such as Dhwani. In terms of ethical considerations, IndicVoices adheres to data protection regulations and obtains informed consent from speakers. This commitment to responsible data collection practices starkly contrasts with approaches that rely on web-mined data, where it is more challenging to adhere to licensing terms. Finally, IndicVoices stands out as the only existing speech corpus that covers all 22 Indian languages, offering rich speaker diversity and a wide range of textual content across various domains and speaking styles. This makes it the ideal candidate for developing a comprehensive TTS dataset.

\subsection{Data Pipeline}


To enhance \dataset{}, we employ a comprehensive data pipeline comprising several key steps which we illustrate in Figure \ref{fig:data-pipeline}.

\textbf{Step 1: Pre-processing audios} IndicVoices consists of recordings at 44.1kHz and 8kHz. We only consider audios at 44.1KHz consisting of extempore and read-speech, as we wish to derive a high-quality speech dataset. We also filter out all audio samples greater than 30s in duration. We upmix mono channels to  stereo channels in all audios using ffmpeg \cite{tomar2006converting}.

\textbf{Step 2: Demixing ASR-quality audios} IndicVoices was recorded in natural environments, which means the audio samples often contain various types of background noise, intermittent or overlapping chatter, echo, and low volume levels. These issues can significantly impact the quality and usability of the dataset for Text-to-Speech (TTS) training. To address these issues, we utilize HTDemucs \cite{defossez2021demucs}, a state-of-the-art deep learning model for audio source separation and noise reduction.

\textbf{Step 3: Dereverberation of Demixed audios} 
Despite the denoiser effectively eliminating most background noises from the audio samples, we encountered instances where the audios still exhibited high reverb and digital artifacts overlapping with the speech or vocals. These artifacts can degrade the overall quality and intelligibility of the audio. By running the entire denoised set through VoiceFixer \cite{liu2021voicefixer}, we observed remarkable improvements in reducing reverberation without any loss of intelligibility in the audio.

\textbf{Step 4: Speech Enhancement} We noticed that VoiceFixer introduced certain digital artifacts in samples. To filter these out, we rely on DeepFilterNet3 \cite{schroeter2023deepfilternet3}, which operates on full-band spectrograms, and found that it effectively eliminates these artifacts while preserving speech quality.

\textbf{Step 5: Filtering} 

\textbf{Audio} We employ a thorough filtering process to ensure high audio quality based on speech quality metrics that we discuss further in Section \ref{subsec:comparison-existing-tts}. Audios must meet specific thresholds: \( \text{C50} \geq \SI{30}{\deci\bel} \) for minimal reverberation, \( \text{SNR} \geq \SI{25}{\deci\bel} \) for clear speech, \( 0.2 < \text{duration} < 30 \) seconds for appropriate length, \( \text{utterance-pitch-mean} \leq \SI{350}{\hertz} \) for natural pitch, \( \text{utterance-pitch-standard-deviation} \leq \SI{150}{\hertz} \) for pitch consistency, and \( \text{speaking-rate} \leq 30 \) characters per second for mainstream pace. This rigorous selection process ensures that our dataset consists of high-quality audio samples appropriate for TTS training, free from irregularities that could impede the performance of TTS models trained on such data.


\textbf{Transcript} IndicVoices provides two levels of transcriptions tailored to diverse needs. In Level 1, transcriptions faithfully reflect spoken language, akin to verbatim transcripts. Level 2 adopts standardized word forms, appealing to both linguistic purists and everyday users. To uphold precision and clarity, we exclude utterances with a character error rate surpassing 5\% between the two levels, ensuring that alignment between transcriptions and speech remains consistent and reliable.

\begin{table}[!t]
\setlength{\tabcolsep}{4pt}
\caption{A comparison of IndicVoices-R with other popular TTS datasets, both English datasets and Indic datasets.}
  \label{tab:speech-quality}
  \centering
\begin{tabular}{@{}lrrrrcccr@{}}

\toprule
\textbf{Dataset} & \textbf{\# Uttr.} & \multicolumn{1}{l}{\textbf{\# Hours}} & \multicolumn{1}{l}{\textbf{\# Spk.}} & \multicolumn{1}{l}{\textbf{\# Lang.}} & \multicolumn{1}{l}{\textbf{N-MOS} ($\uparrow$)} & \multicolumn{1}{l}{\textbf{SNR} ($\uparrow$)} & \multicolumn{1}{l}{\textbf{C50} ($\uparrow$)} & \multicolumn{1}{c}{\textbf{F0} ($\uparrow$)} \\ \midrule
LJSpeech & 13.1K & 24 & 1 & 1 & 4.36 & 58.03 & 58.56 & 208.04 \\
LibriTTS & 154.6K & 585 & 2456 & 1 & 3.76 & 59.60 & 57.33 & \cellcolor[HTML]{FFFFFF}159.51 \\
IndicTTS & 141.9K & 284 & 27 & 14 & 4.29 & \cellcolor[HTML]{FFFFFF}65.38 & 58.78 & 166.04 \\
Google-CS & 25.2K & 41 & 261 & 7 & 4.20 & \cellcolor[HTML]{FFFFFF}61.63 & 53.24 & 192.86 \\
LIMMITS & 246K & 560 & 14 & 7 & 4.30 & 58.69 & 59.60 & 176.37 \\
\midrule
\textbf{Ours} & 689.6K & 1704 & 10496 & 22 & 3.38 & 60.47 & 53.45 & 178.91 \\ \bottomrule
\end{tabular}
\end{table}

\textbf{Step 6: Post-Processing} We randomly sampled audio files from the enhanced set and found that they differed in volume. To fix this, we employed the normalize function in PyDub \cite{robert2018pydub} to adjust the volume of an audio segment and bring its peak amplitude close to the maximum possible level, while maintaining a specified headroom (set to 0.1 in dB) to avoid clipping. Just as we ensure consistency in audio volume through normalization, one might question the value of normalizing text. Since we use the verbatim version of the transcript, no additional normalization is required and we are able to maintain fidelity to a speaker's utterances, capturing nuances, colloquialisms, and idiosyncrasies inherent in natural language.


\subsection{Comparison with Existing TTS Datasets}
\label{subsec:comparison-existing-tts}
 To prove the utility of our data pipeline in yielding high-quality samples, we compare \dataset{} against popular TTS datasets including LJSpeech, LibriTTS, IndicTTS, Google-CS and LIMMITS. We attempt to show that the speech quality of our dataset is on par with other existing TTS datasets while surpassing them in terms of speaker diversity, vocabulary diversity, and corpus size. 


\subsubsection{Speech Quality}
\begin{figure}
  \centering
  \begin{minipage}[!t]{0.45\linewidth}
    \small
    \centering
    \captionof{table}{Distribution of number of speakers (S) and duration in hours (H) of our data across age groups and genders.}
    \begin{tabular}{@{}ccccc@{}}
    \toprule
    \textbf{Age}   & \multicolumn{2}{c}{\textbf{Male}} & \multicolumn{2}{c}{\textbf{Female}} \\ \midrule
    \textbf{Group} & \textbf{\# S}   & \textbf{\# H}   & \textbf{\# S}    & \textbf{\# H}    \\ \midrule
    18-30          & 2109            & 341.69          & 2284             & 369.72           \\
    30-45          & 1414            & 234.87          & 1575             & 250.85           \\
    45-60          & 881             & 154.04          & 1015             & 162.06           \\
    60+            & 626             & 108.80          & 592              & 97.18            \\
    \midrule
    \textbf{Total} & 5030            & 839.41          & 5466             & 879.80           \\ \bottomrule
    \end{tabular}
    \label{tab:age-group-distribution}
  \end{minipage}
  \hfill
  \begin{minipage}[!t]{0.45\linewidth}
    \centering
    \includegraphics[width=\linewidth]{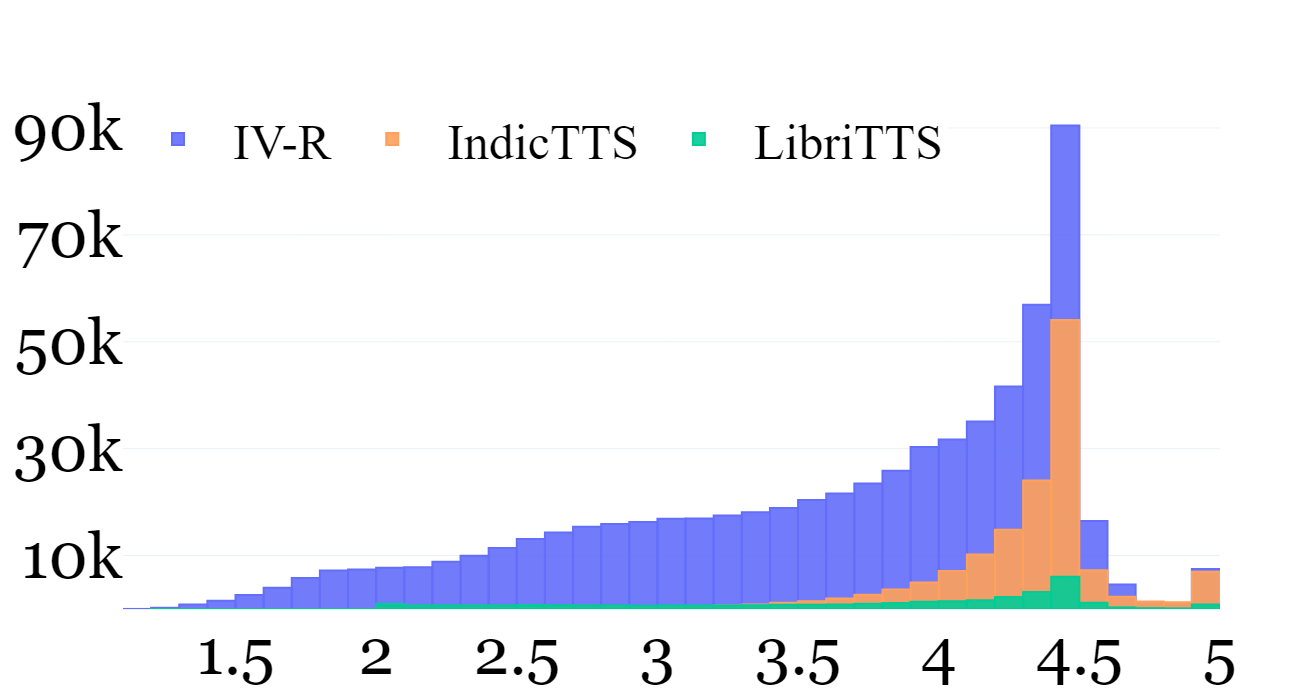}
    \captionof{figure}{N-MOS of IV-R vs. existing TTS datasets.}
    \label{fig:nmos}
  \end{minipage}
\end{figure}

\begin{figure}
  \begin{minipage}[b]{0.5\linewidth}
    \centering
     \includegraphics[width=\linewidth]{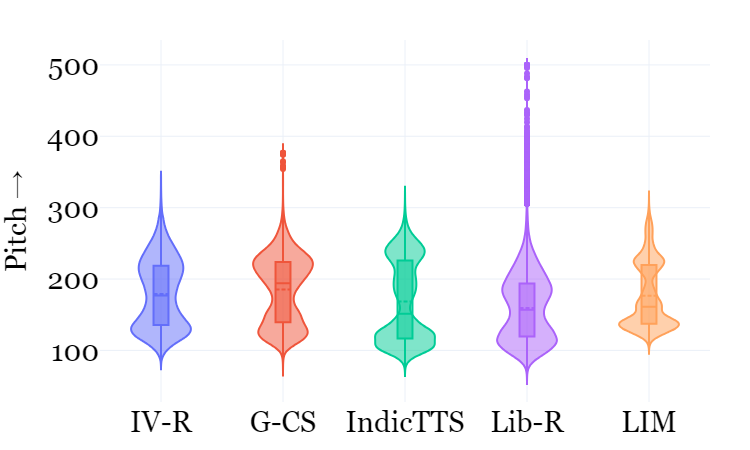}
    \label{fig:pitch-distribution}
    \end{minipage}
  \begin{minipage}[b]{0.5\linewidth}
    \centering
    \includegraphics[width=\linewidth]{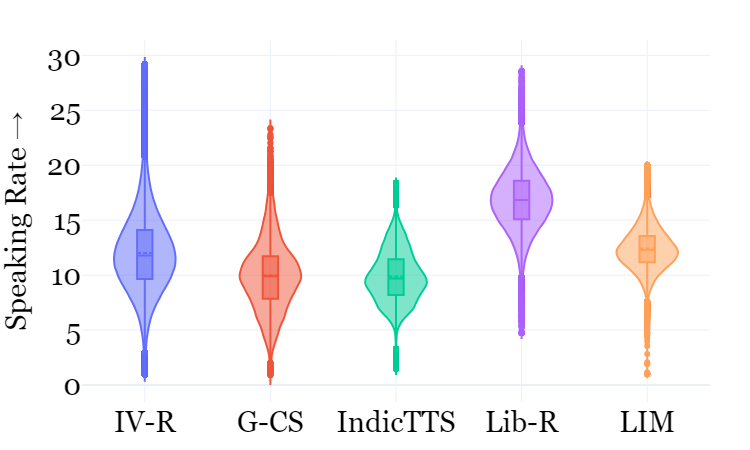}
    \label{fig:speaking-rate-distribution}
  \end{minipage}
  \caption{Comparison of Pitch and Speaking Rates of IV-R vs. existing TTS datasets.}
  \label{speaker_stats}
\end{figure}
We report four metrics (i) NORESQA-MOS(N-MOS), (ii) Signal-to-Noise Ratio (SNR), (iii) C50 and (iv) utterance-level pitch mean ($F_0$) values to measure the quality and clarity of the speech samples. The NORESQA \cite{noresqamos} framework uses a Non-Matching Reference (NMR) along with the given test speech signal to estimate speech quality. Particularly, we calculate the N-MOS score using randomly chosen samples from the LibriTTS set as the NMR. We used Brouhaha \cite{lavechin2023brouhaha} to compute SNR and C50 and PENN \cite{morrison2023cross} to compute utterance level pitch mean ($F_0$) using the Dataspeech \cite{lacombe-etal-2024-dataspeech} repository.
In Table \ref{tab:speech-quality}, we highlight that \dataset{} surpasses all other datasets in terms of sheer volume of number of utterances, hours, speakers, and languages. Furthermore, while studio-quality TTS datasets such as LJSpeech, IndicTTS, and LIMMITS achieve higher N-MOS scores than ASR-purposed datasets like LibriTTS and IndicVoices-R. Nevertheless, we plot the distribution of N-MOS (Figure \ref{fig:nmos}) across datasets and find that \dataset{} has more utterances than IndicTTS and LibriTTS with N-MOS scores greater than 4. 

Our dataset also maintains competitive noise levels (measured by SNR) with other datasets, indicating its suitability for TTS. Its C50 score of \SI{53.45}{\deci\bel}, is only slightly lower than studio-quality datasets (LJSpeech, IndicTTS, LIMMITS), indicating good speech clarity. In contrast, studio-quality datasets like LJSpeech, IndicTTS, and LIMMITS have higher mean C50 values, reflecting the superiority of recording in studio environments. The $F_0$ values, representing pitch, are within a comparable range across other Indian TTS datasets, indicating expected pitch levels. Overall, \dataset{}, with its extensive linguistic diversity and considerable dataset size, offers a balance between quality and quantity, making it a valuable dataset for building TTS systems.

\subsubsection{Speech  Diversity}

\textbf{Speaker Diversity} Covering as many as 10,469 speakers across the 22 official Indian languages, \dataset{} has 40$\times$ more speakers than the previous Indian TTS corpora with the highest speaker diversity, Google-CS. 
Table \ref{tab:age-group-distribution} elucidates the diversity and balance in the distribution of the number of speakers and data hours in each age group. This comprehensive diversity is pivotal for training TTS systems capable of zero-shot speaker and style adaptation.

\textbf{Style Diversity}: The analysis of pitch distribution and speaking rates across various datasets, as depicted in Figure \ref{speaker_stats}, underscores the inherent naturalness of our \dataset{}. Our dataset exhibits a rich array of mean pitch values, extending to 350 Hz, contrasting with the narrower pitch range observed in datasets like IndicTTS, which predominantly comprises meticulously recorded read-speech in controlled settings. Furthermore, our dataset demonstrates a mean speaking rate of 12 words per second, with a notable dispersion, in contrast to the more uniform rate of 9.88 words per second found in IndicTTS, indicating a prevalence of spontaneous speech and naturalness.

\begin{figure}
    \includegraphics[width=\linewidth]{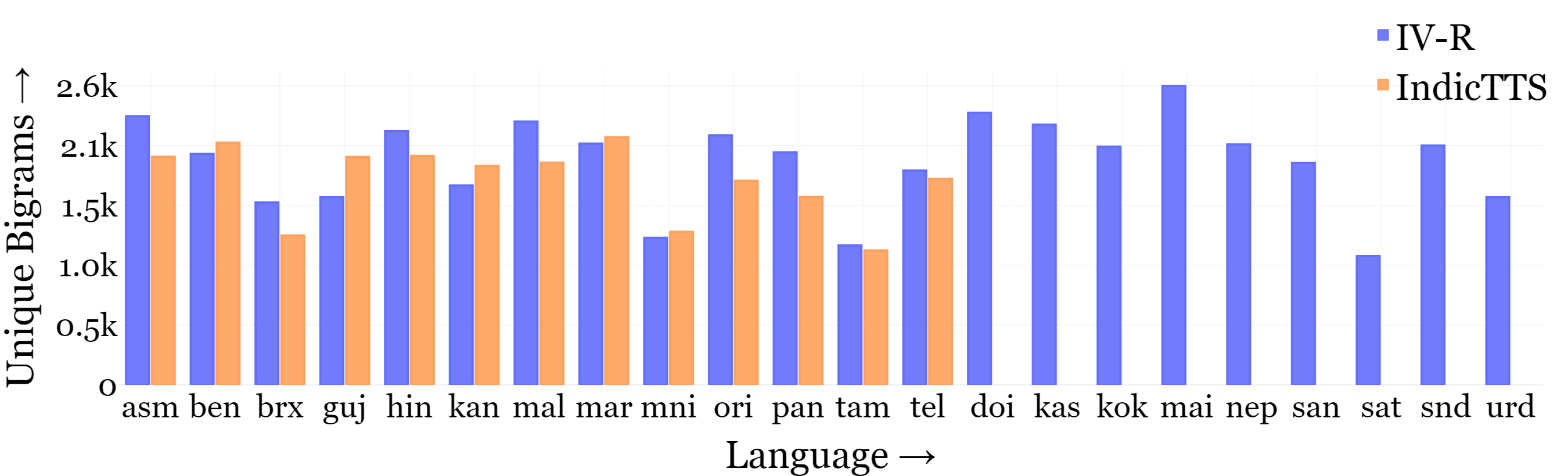}
    \label{bigram}
    \caption{Unique character bigram counts across languages for our dataset and IndicTTS. Languages are reported using ISO-639-1 language codes.
}
\label{bigrams-plot}
\end{figure}

\textbf{Vocabulary Diversity}
We compare character bigram for \dataset{} and IndicTTS in Figure \ref{bigrams-plot} and found our data to be marginally better in 8 languages while comparable in the remaining 5 languages that IndicTTS covers.

\begin{table}
\caption{Detailed Statistics of \dataset{} across languages}
\label{tab:detailed-statistics}
\centering
\setlength{\tabcolsep}{3pt} 
\renewcommand{\arraystretch}{1} 
\begin{tabular}{@{}lrrrrrrrcc@{}}
\toprule
\textbf{Languages} & \multicolumn{3}{c}{\textbf{\# Hours}}                                                                           & \multicolumn{2}{c}{\textbf{\# Utterances}}                                 & \multicolumn{2}{c}{\textbf{\# Speakers}}                                   & \textbf{\# Words}    & \textbf{\# Bi-grams} \\ \cmidrule(lr){2-4}\cmidrule(lr){5-6}\cmidrule(lr){7-8}
                   & \multicolumn{1}{c}{\textbf{Read}} & \multicolumn{1}{c}{\textbf{Extempore}} & \multicolumn{1}{c}{\textbf{Total}} & \multicolumn{1}{c}{\textbf{Count}} & \multicolumn{1}{c}{\textbf{Avg. (s)}} & \multicolumn{1}{c}{\textbf{Count}} & \multicolumn{1}{c}{\textbf{Avg. (s)}} & \multicolumn{1}{l}{} & \multicolumn{1}{l}{} \\
                   \midrule
\textbf{Assamese}  & 7.05                              & 168.29                                 & \cellcolor[HTML]{A4C2F4}175.34     & 73077                              & 8.64                                  & \cellcolor[HTML]{FEAC33}928        & 680.43                                & 73846                & 2302                 \\
Bengali            & 4.63                              & 107.37                                 & \cellcolor[HTML]{C0D7F3}111.99     & 40943                              & 9.85                                  & \cellcolor[HTML]{FCD298}617        & 654.20                                & 45095                & 1979                 \\
Bodo               & 2.17                              & 169.89                                 & \cellcolor[HTML]{A6C4F3}172.05     & 83976                              & 7.38                                  & \cellcolor[HTML]{FEAB2F}941        & 658.90                                & 97526                & 1563                 \\
Dogri *            & 3.64                              & 67.04                                  & \cellcolor[HTML]{D2E4F4}70.68      & 27967                              & 9.10                                  & \cellcolor[HTML]{FCE3C7}470        & 541.16                                & 33127                & 2330                 \\
Gujarati           & 1.06                              & 7.88                                   & \cellcolor[HTML]{FFFFFF}8.94       & 3304                               & 9.74                                  & \cellcolor[HTML]{FFFFFF}45         & 707.91                                & 11773                & 1607                 \\
Hindi              & 4.31                              & 70.28                                  & \cellcolor[HTML]{D0E3F4}74.60      & 27557                              & 9.75                                  & \cellcolor[HTML]{FDE9D4}399        & 672.83                                & 24697                & 2173                 \\
Kannada            & 3.87                              & 40.74                                  & \cellcolor[HTML]{E5F0F9}44.61      & 18127                              & 8.86                                  & \cellcolor[HTML]{FCE5CD}452        & 354.56                                & 53683                & 1708                 \\
Kashmiri *         & 7.12                              & 57.87                                  & \cellcolor[HTML]{D7E7F5}64.99      & 26134                              & 8.95                                  & \cellcolor[HTML]{FDE6CE}450        & 517.89                                & 50424                & 2230                 \\
Konkani  *          & 5.87                              & 47.19                                  & \cellcolor[HTML]{DFECF7}53.06      & 22357                              & 8.54                                  & \cellcolor[HTML]{FEF4E9}228        & 839.34                                & 47467                & 2041                 \\
Maithili *         & 6.18                              & 75.59                                  & \cellcolor[HTML]{CDE0F3}81.77      & 32483                              & 9.06                                  & \cellcolor[HTML]{FCD095}627        & 462.78                                & 50560                & 2561                 \\
Malayalam          & 5.47                              & 77.10                                  & \cellcolor[HTML]{CCE0F3}82.57      & 32544                              & 9.13                                  & \cellcolor[HTML]{FCE4CA}462        & 641.73                                & 90052                & 2256                 \\
Manipuri           & 0.77                              & 23.22                                  & \cellcolor[HTML]{F5F9FD}23.99      & 9312                               & 9.28                                  & \cellcolor[HTML]{FFFAF5}127        & 664.05                                & 32352                & 1259                 \\
Marathi            & 4.80                              & 46.10                                  & \cellcolor[HTML]{E1EDF8}50.90      & 20164                              & 9.09                                  & \cellcolor[HTML]{FDEBD9}359        & 507.50                                & 43059                & 2066                 \\
Nepali *           & 8.89                              & 96.99                                  & \cellcolor[HTML]{C2D9F3}105.87     & 43545                              & 8.75                                  & \cellcolor[HTML]{FDC678}716        & 533.93                                & 58952                & 2060                 \\
Odia               & 6.04                              & 64.90                                  & \cellcolor[HTML]{D2E4F4}70.95      & 26450                              & 9.66                                  & \cellcolor[HTML]{FDE6CF}441        & 579.28                                & 37601                & 2138                 \\
Punjabi            & 6.20                              & 68.74                                  & \cellcolor[HTML]{CFE2F3}74.94      & 27788                              & 9.71                                  & \cellcolor[HTML]{FDEDDC}335        & 805.40                                & 25910                & 1992                 \\
Sanskrit *         & 4.82                              & 30.93                                  & \cellcolor[HTML]{ECF4FB}35.75      & 14604                              & 8.81                                  & \cellcolor[HTML]{FFF8F1}161        & 787.51                                & 35271                & 1901                 \\
Santali *          & 6.31                              & 70.06                                  & \cellcolor[HTML]{CFE2F3}76.37      & 35155                              & 7.82                                  & \cellcolor[HTML]{FEEFDF}309        & 890.28                                & 33102                & 1104                 \\
Sindhi *           & 2.50                              & 7.98                                   & \cellcolor[HTML]{FEFFFF}10.48      & 4197                               & 8.99                                  & \cellcolor[HTML]{FEF5EC}204        & 191.67                                & 11900                & 2050                 \\
Tamil              & 11.94                             & 87.53                                  & \cellcolor[HTML]{C5DBF3}99.47      & 40464                              & 8.85                                  & \cellcolor[HTML]{FF9900}1084       & 328.37                                & 96167                & 1195                 \\
Telugu             & 6.91                              & 129.49                                 & \cellcolor[HTML]{B5CFF3}136.40     & 48485                              & 10.13                                 & \cellcolor[HTML]{FDCA83}681        & 721.46                                & 90301                & 1837                 \\
Urdu *             & 4.45                              & 74.17                                  & \cellcolor[HTML]{CEE1F3}78.61      & 30935                              & 9.15                                  & \cellcolor[HTML]{FCE4CB}460        & 624.67                                & 23752                & 1607                 \\ \midrule
\textbf{Total}     & 114.98                            & 1,589.36                               & \textbf{1,704.34}                  & 689568                             & \multicolumn{1}{c}{-}                 & \textbf{10496}                     & \multicolumn{1}{c}{-}                 & 1066617              & 41959                \\ \bottomrule
\end{tabular}
\end{table}



\subsection{Data Statistics and Format}

We present comprehensive statistics of \dataset{} in Table \ref{tab:detailed-statistics}. Notably, our dataset is the first to publicly open-source TTS data for 9 Indian languages - Dogri, Kashmiri, Konkani, Maithili, Nepali, Sanskrit, Santali, Sindhi, and Urdu. Overall, our dataset consists of 1,704 hours of high-quality speech data from 10,496 speakers across 22 Indian languages with over 1M unique words and 690K utterances. We release the metadata files in JSONL format and explain the attributes provided for each speech-text pair in Appendix.

\section{IndicVoices-R Benchmark}
\label{sec:benchmark}
We complement \dataset{} by also providing a train set and a carefully crafted held-out set to truly test the zero-shot, few-shot, and many-shot speaker generalization capabilities of TTS systems for Indian speakers and languages. We call this benchmark as \dataset{} Benchmark and illustrate its statistics in Table \ref{tab:benchmark-detailed-statistics}. To construct this benchmark, we attempt to maximize the number of zero-shot and few-shot speakers across both genders and age groups while also ensuring we do not lose out on too much training data. We select the bottom 2 speakers per gender age-group pair with the least durations and include them in the zero-shot test set.  For few-shot and many-shot splits,  we uniformly sample utterances across genders, age groups, durations, and pitch in an attempt to ensure style diversity. Thus, this benchmark serves as a novel evaluation framework encompassing age and gender diversity to uniquely assess TTS performance across various demographics, ensuring robust generalization of models.

\begin{table}[H]
\caption{Speaker counts and test durations of \dataset{} Benchmark expanded across genders, age groups, and N-shot splits where ZS implies zero-shot (no training data for speaker), and $<K$ mins implies the speaker has less than $K$ minutes in the training data.}
\label{tab:benchmark-detailed-statistics}
\setlength{\tabcolsep}{3pt} 
\renewcommand{\arraystretch}{1} 
\centering
\begin{tabular}{llrrrrrrrr}
\toprule \textbf{Gender} & \multicolumn{1}{c}{\textbf{Age}} & \multicolumn{4}{c}{\textbf{Speakers}} & \multicolumn{4}{c}{\textbf{Durations (in hours)}}          \\
\cmidrule(lr){3-6}\cmidrule(lr){7-10}
       & \multicolumn{1}{c}{\textbf{Group}} & \multicolumn{1}{c}{\textbf{ZS}} & \multicolumn{1}{c}{\textbf{\textless{}5 mins}} & \multicolumn{1}{c}{\textbf{\textless{}10 mins}} & \multicolumn{1}{c}{\textbf{\textgreater{}10 mins}} & \multicolumn{1}{c}{\textbf{ZS}}  & \multicolumn{1}{c}{\textbf{\textless 5 mins}} & \multicolumn{1}{c}{\textbf{\textless{}10 mins}} & \multicolumn{1}{c}{\textbf{\textgreater{}10 mins}} \\ \midrule
Male   & 18-30      & 45                           & 107                                   & 272                                    & 468                          & 1.32                          & 0.08                                 & 2.56                                   & 0.75                                      \\
       & 30-45      & 44                           & 75                                    & 177                                    & 308                          & 1.72                          & 0.06                                 & 2.13                                   & 0.95                                      \\
       & 45-60      & 44                           & 41                                    & 130                                    & 233                          & 1.98                          & 0.03                                 & 2.44                                   & 0.56                                      \\
       & 60+        & 44                           & 29                                    & 87                                     & 159                          & 2.85                          & 0.02                                 & 1.57                                   & 1.25                                      \\ \midrule
Female & 18-30      & 46                           & 130                                   & 258                                    & 409                          & 1.57                          & 0.08                                 & 2.23                                   & 1.08                                      \\
       & 30-45      & 44                           & 75                                    & 187                                    & 280                          & 1.51                          & 0.06                                 & 2.26                                   & 0.54                                      \\
       & 45-60      & 44                           & 59                                    & 135                                    & 178                          & 2.21                          & 0.05                                 & 2.03                                   & 0.92                                      \\
       & 60+        & 41                           & 25                                    & 78                                     & 91                           & 2.41                          & 0.03                                 & 1.66                                   & 0.58                                      \\ \midrule
\textbf{Total}       &            & 352                          & 541                                   & 1324                                   & 2126                         & 15.55                         & 0.41                                 & 16.88                                  & 6.63                                     \\ \bottomrule
\end{tabular}
\end{table}

\section{Experiments with IndicVoices-R}
\label{sec:Training-details}


We employ VoiceCraft \cite{peng2024voicecraft}, a Transformer-based architecture that employs a novel token re-arrangement and delayed stacking mechanism to causally predict audio codec tokens. Particularly, to study the utility of our dataset, we conduct two fine-tuning experiments starting from the publicly available  830M parameter checkpoint \footnote{Pre-trained checkpoint: \url{https://huggingface.co/pyp1/VoiceCraft/blob/main/giga830M.pth}} trained on GigaSpeech. Specifically, we first fine-tune a model on the IndicTTS database and compare its performance on the benchmark against a model fine-tuned on \dataset{}.

\textbf{Vocabulary Expansion} To facilitate the fine-tuning of the English pre-trained model on Indian languages, we extend the tokenizer and initialize the newly introduced token embeddings with a Gaussian distribution centered around $\mu_{old}$ with variance $\sigma^2_{old}$. The token embeddings are concatenated with the old embeddings. The vocab size was extended from 100 tokens to 1089 tokens pooled across all 22 languages. The tokenizer was extended to 1536 with a token padding multiple of 512 for training efficiency. We chose graphemes over phonemes for building our multilingual TTS system due to the lack of phonemizers for all Indian languages.




\textbf{Experimental Setup} We train VoiceCraft on 8x NVIDIA A100 40GB GPUs. We use the standard AdamW optimizer with $\beta_1=0.99$ and $\beta_2=0.999$, a weight decay of $\lambda=10^{-2}$ and finetune it for over 50K steps. We use the Lambda learning rate schedule with an initial learning rate of $0.00001$, and dynamically batched with a maximum of 20,000 tokens per GPU.

\textbf{Results} We report N-MOS and speaker similarity (S-SIM) scores. To calculate speaker similarity, we present the cosine similarity between the embeddings of the ground-truth and synthesized samples, extracted from Wav2vec2 \cite{wav2vec2} fine-tuned on IV-R data using the IndicSUPERB pipeline \cite{superb2023javed}. As observed in Table \ref{tab:finetune-results}, VoiceCraft fine-tuned on \dataset{} achieves better speaker similarity scores on the zero-shot benchmark with nearly on-par N-MOS compared to the model fine-tuned on IndicTTS.


\begin{table}[H]
\centering
\caption{Zero-shot speaker evaluation of VoiceCraft, fine-tuned on IndicTTS and dataset, on the \dataset{} Benchmark.}
\label{tab:finetune-results}
\begin{tabular}{@{}lrr@{}}
\toprule
\textbf{Finetune dataset} & \textbf{NORESQA} & \textbf{S-SIM} \\ \midrule
IndicTTS & 3.83 & 78.93 \\   
\textbf{IndicVoices-R (Ours)} & 3.64 & \textbf{88.18} \\ \bottomrule  
\end{tabular}
\end{table}

\section{Ethical Considerations and Limitations}
\label{sec:Ethics}
We prioritize responsible data practices in developing IndicVoices-R (IV-R). The dataset leverages anonymized speech from its parent dataset IndicVoices and is released under the same CC BY 4.0 license. IndicVoices underwent rigorous ethical review and approval by the Institute Ethics Committee. It obtained explicit consent from each participant for the use of their speech data and ensured no offensive content was part of it. We acknowledge potential biases within the source data and encourage further exploration. Addressing the limitations of our work, we could not unlock the potential of conversational subset of IndicVoices due to their low sampling rates of \SI{8}{\kilo\hertz}. Furthermore, although the dataset exhibits characteristics of natural speech, it has not attained the caliber of studio-recorded speech datasets. Henceforth, our commitment persists in the pursuit of continual enhancement and advancement.

\section{Conclusion}
\label{sec:conclusion}
In this work, we introduce IndicVoices-R, a first-of-its-kind large-scale multilingual TTS dataset encompassing 22 Indian languages. This dataset comprises 1704 hours of high-fidelity speech-text data, capturing a diverse range of scenarios and speaker demographics. IndicVoices-R facilitates research and development in zero-shot multilingual text-to-speech synthesis. To complement the dataset, we further introduce a comprehensive benchmark designed to evaluate the zero-shot, few-shot, and many-shot capabilities of TTS models on data across various age and gender groups, along with different text and speech styles. We empirically demonstrate the dataset's utility by training a competent zero-shot TTS system on IndicVoices-R. Our work paves the way for the development of more robust and speaker-agnostic TTS systems, ultimately fostering broader accessibility and inclusivity for Indian languages.
% ...

\bibliographystyle{IEEEtran}
\bibliography{refs, rasa}
\label{sec:references}

\end{document}